\documentclass[conference]{IEEEtran}
\IEEEoverridecommandlockouts
\usepackage{cite}
\usepackage{amsmath,amssymb,amsfonts}
\usepackage{algorithmic}
\usepackage{graphicx}
\usepackage{textcomp}
\usepackage{xcolor}

\newcommand{\Real}{\mathbb{R}}
\newcommand{\Natural}{\mathbb{N}}

\newtheorem{thm}{Theorem}
\newtheorem{rem}{Remark}

\newtheorem{lem}{Lemma}

\newtheorem{assume}{Assumption}
\newtheorem{problem}{Problem}

\newcommand{\bfx}{\boldsymbol{x}}
\newcommand{\bfu}{\boldsymbol{u}}

\newcommand{\bfz}{\boldsymbol{z}}
\newcommand{\bfc}{\boldsymbol{c}}

\newcommand{\bfy}{\boldsymbol{y}}

\def\BibTeX{{\rm B\kern-.05em{\sc i\kern-.025em b}\kern-.08em
    T\kern-.1667em\lower.7ex\hbox{E}\kern-.125emX}}
\begin{document}

\title{Demystification of Few-shot and One-shot Learning\\
\thanks{The work was supported by the UKRI Alan Turing AI Acceleration Fellowship grant EP/V025295/1 and by the grant of the Ministry of Science and Higher Education of Russian Federation (Project No. 14.Y26.31.0022).}
}

\author{\IEEEauthorblockN{1\textsuperscript{st} Ivan Y. Tyukin}
\IEEEauthorblockA{\textit{School of Mathematics}\\ 
\textit{and Actuarial Science} \\
\textit{University of Leicester}\\
and \textit{Norwegian University}\\
\textit{of Science and Technology}\\
and \textit{Saint-Petersburg State}\\
\textit{Electrotechnical University}\\
\\
Leicester, United Kingdom \\
and Trondheim, Norway\\
and Saint-Petersburg, Russia\\
i.tyukin@le.ac.uk}
\and
\IEEEauthorblockN{2\textsuperscript{nd} Alexander N. Gorban}
\IEEEauthorblockA{\textit{School of Mathematics}\\
\textit{and Actuarial Science} \\
\textit{University of Leicester}\\
and \textit{Lobachevsky University}\\
\\
Leicester, United Kingdom \\
and Nizhny Novgorod, Russia\\
a.n.gorban@le.ac.uk}
\and
\IEEEauthorblockN{3\textsuperscript{rd} Muhammad H. Alkhudaydi}
\IEEEauthorblockA{\textit{School of Mathematics}\\
\textit{and Actuarial Science} \\
\textit{University of Leicester}\\
\\
Leicester, United Kingdom \\
mhaa4@le.ac.uk}
\and
\IEEEauthorblockN{4\textsuperscript{th} Qinghua Zhou}
\IEEEauthorblockA{\textit{School of Informatics} \\
\textit{University of Leicester}\\
\\
Leicester, United Kingdom \\
qz105@le.ac.uk}
}

\maketitle

\begin{abstract}
Few-shot and one-shot learning have been the subject of active and intensive research in recent years, with mounting evidence pointing to successful implementation and exploitation of few-shot learning algorithms in practice. Classical statistical learning theories do not fully explain why few- or  one-shot learning is at all possible since traditional generalisation bounds normally require large training and testing samples to be meaningful. This sharply contrasts with numerous examples of successful one- and few-shot learning systems and applications. 

In this work we present mathematical foundations for a theory of one-shot and few-shot learning and reveal conditions specifying when such learning schemes are likely to succeed. Our theory is based on intrinsic properties of high-dimensional spaces. We show that if the ambient or latent decision space of a learning machine is sufficiently high-dimensional than a large class of objects in this space can indeed be easily learned from few examples provided that certain data non-concentration conditions are met.
\end{abstract}


\section*{Notation}

\begin{itemize}
	\item {$\Real$ denotes the field of real numbers, $\Real_{\geq 0}=\{x\in\Real| \ x\geq 0\}$, and} $\Real^n$ stands for the $n$-dimensional linear real vector space;
	\item $\Natural$ denotes the set of natural numbers;
	\item bold symbols $\boldsymbol{x} =(x_{1},\dots,x_{n})$ will denote elements of $\Real^n$;
	\item $(\boldsymbol{x},\boldsymbol{y})=\sum_{k} x_{k} y_{k}$ is the inner product of $\boldsymbol{x}$ and $\boldsymbol{y}$, and $\|\boldsymbol{x}\|=\sqrt{(\boldsymbol{x},\boldsymbol{x})}$ is the standard Euclidean norm  in $\Real^n$;
	\item  $\mathbb{B}_n$ denotes the unit ball in $\Real^n$ centered at the origin:
	\[\mathbb{B}_n=\{\boldsymbol{x}\in\Real^n| \ {\|\boldsymbol{x}\|\leq 1}\};\]
	\item  $\mathbb{B}_n(r,\bfy)$  stands for the ball in $\Real^n$ of radius ${r> 0}$ centered at $\bfy$: 
	\[\mathbb{B}_n(r,\bfy)=\{\boldsymbol{x}\in\Real^n| \ {\|\boldsymbol{x}-\bfy\|\leq r}\};\]
	\item $V_n$ is the $n$-dimensional Lebesgue measure, and $V_n(\mathbb{B}_n)$ is the volume of unit {$n$}-ball;
\end{itemize}

\section{Introduction}

The fundamental question of learning from few examples is one of the fascinating and  central  questions in both the theory and practice of  modern large-scale data-driven AI systems. These systems have many millions of adjustable parameters \cite{sandler2018mobilenetv2}, whose numbers often exceed those of the datasets used in their training. And yet, performance of these large-scaled models trained on  modestly-sized datasets in practical tasks is remarkably good \cite{parkhi2015deep}.

Classical generalisation bounds stemming from the Vapnik-Chervonenkis theory \cite{vapnik1999overview} alone do not explain these successes due to their combinatorial and extremely conservative nature. What is even more striking, as  has been demonstrated in \cite{zhang2016understanding}, absolutely identical deep neural networks are capable of exhibiting both sides of the learning spectrum: to successfully generalise from meaningful training data and, at the same time, ``memorise'' random assignments of labels without any generalisation. Results like these motivate persistent ongoing inquiries into unreasonable effectiveness of modern deep learning models \cite{sejnowski2020unreasonable}.

The phenomenon of few-shot learning is perhaps one of the most acute manifestations of this challenge. Various few-shot learning schemes and empirically successful algorithms and models such as  matching \cite{vinyals2016matching} and prototypical networks \cite{snell2017prototypical} provide ample evidence that good generalisation may indeed occur in extreme settings with just few elements in the training set. The theory, however, which may explain why is this at all  possible is lacking.

In this paper, we lay out mathematical foundations of such theory. We provide, for the first time, {\it formal statements} of different versions of the problem of few-shot learning and {\it present solutions} of these problems. These solutions are remarkably consistent with heuristic algorithms described in the current literature \cite{vinyals2016matching}, \cite{snell2017prototypical}. At the core of our  approach are stochastic separation theorems \cite{GorbanTyukin:NN:2017}, \cite{grechuk2020general} linking high-dimensional geometry with the concentration of measure. In this work, we make an additional departure from the classical ``fully agnostic'' machine learning problem statement. In particular, we propose that a mild hypothesis on ``compactness'' of an object's/class's representation in the network's latent space, expressed as existence of a finite sub-cover of the object to be learned by $n$-balls not containing the origin, could hold the key to understanding and resolving the challenge of generalisation, few-shot, and single-shot learning. 

The rest of the paper is organised as follows. In Section \ref{sec:problem_statement} we describe a general setting of the problem  of few-shot learning considered in the paper  and present its formal mathematical statements. Section \ref{sec:results} presents main mathematical results and their discussion, and Section \ref{sec:conclusion} provides a brief summary and conclusion.

\section{Problem formulation}\label{sec:problem_statement}

\subsection{General setting}

To set the scene for a more formal analysis, let us first outline key components of few-shot learning. In many relevant few-shot learning cases one would normally have an {\it existing system} with all its inputs, outputs, states and dependencies (potentially unknown) between these. This existing system would also operate in a specific regime (recognise a new person in a room, learn  a new gesture, fix an error) which can be termed as an {\it operational situation}. Performance of the system in the task of  learning in this situation is then assessed by some  {\it evaluation procedure}.  

Complexity of all processes presented in this rather generic picture could be extremely high.  In modern large-scale AI and deep learning models, one of the major contributors to this complexity is an inherently and irreducibly {\it high dimensionality} of signals involved in the definition of the operational situation at hand. Defining meaningful probability spaces for such data is not a trivial task due to enormously large datasets required to gain appropriate knowledge and intuition. At the same time, as we will show later, this high dimensionality may hold the key to develop some understanding of the phenomenon of few-shot learning.

In order to reveal the link between dimensionality of the appropriate data and few-shot learning we will need to make some simplifying assumptions constraining the general setting above. These assumptions, however, would enable us to define the problem formally and focus on the most relevant elements of the general problem which are important for this contribution. In the next section we provide a formal, albeit simplified, description of the problem (\ref{sec:background}), formalise the problem of few-shot learning (\ref{sec:problem}), and list some specific  technical assumptions (\ref{sec:assumptions}).


\subsection{Background}\label{sec:background}

Let $\mathcal{U}$, $\mathcal{U}\subset\Real^d$ be the set of inputs modeling or representing objects of  interests such as images, pieces of sound, or records in a database, and let $\mathcal{L}$ be the set of labels. Following classical statistical learning settings \cite{cucker2002mathematical}, \cite{vapnik1999overview}, we suppose that for each element $\bfu\in\mathcal{U}$  there is an associated  label $\ell\in\mathcal{L}$, and that the pairs $(\bfu_i,\ell_i)\in\mathcal{U}\times\mathcal{L}$, $i=1,\dots,N$, $N\in\Natural$ are
observations drawn from some joint probability distribution.  For convenience, we shall assume that there exist some corresponding distributions  $P_u$ and $P_{\ell|u}$ such that the joint distribution of $\bfu$ and $\ell$ is expressed as: $P_{\ell|u}(\ell|\bfu)P_u(\bfu)$. 

To formally specify the problem of few-shot learning and its relevant variants, we need to determine a system that would be subjected to such learning. For the sake of simplicity, here we will assume that this system is a classifier. In general, however, this latter assumption may be dropped, and the problem of few-shot learning could be extended to much broader classes of AI systems.

\begin{figure}
\centering
\includegraphics[width=0.8\columnwidth]{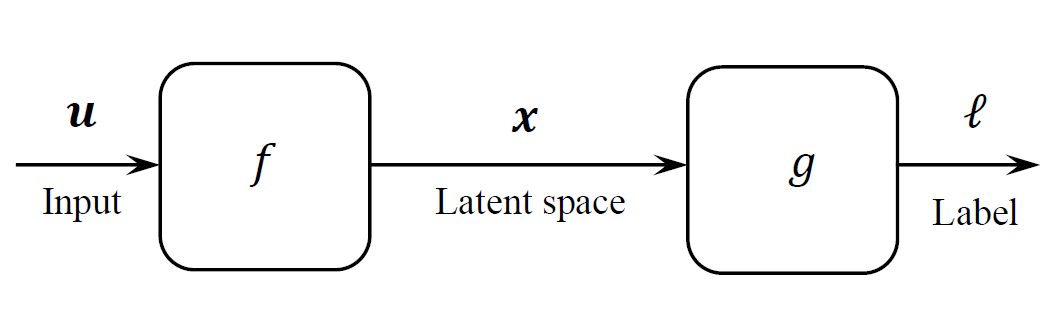}
\caption{Assumed input-output workflow of the classifier subject to few-shot learning tasks}\label{fig:scheme_classifier}
\end{figure}

Let $F:\mathcal{U}\rightarrow\mathcal{L}$ be such classifier assigning a unique label from the set $\mathcal{L}$ (the set defining all possible lables) to an element from $\mathcal{U}$. In what follows we shall assume that $F=g\circ f$ (and {\it denoted} as $F[g\circ f]$ ) where
\begin{equation}\label{eq:classifier_general}
f: \mathcal{U}\rightarrow \mathcal{X}, \ \mathcal{X}\subset \Real^n
\end{equation}
defines the classifier's $F$ latent space $\mathcal{X}$, and
\[
g: \mathcal{X}\rightarrow \mathcal{L} 
\]
determines how the classifier $F$ assigns a label to an input $\bfu$  having the the corresponding latent representation $\bfx=f(\bfu)$. A diagram showing schematic representation of the classifier's workflow is shown in Fig. \ref{fig:scheme_classifier}. 

The above structure is very general and covers the majority of existing classification models.  We are now ready for formal definitions of the relevant few-shot learning problems.

\subsection{Few-shot learning problems}\label{sec:problem}

In what follows we consider two classes of few-shot learning problems: {\it learning new examples} from their single representation, and {\it learning a new class} from few examples. These problems have different uses and aims. The latter focuses primarily on {\it generalising} from a limited number of data points, whereas the former aims at {\it memorising} new data without destroying existing knowledge in the system. 

\subsubsection{Learning a finite number of new examples}

We begin with the first version of the problem, where the task is to learn, or memorise, a given finite set. This task is formally introduced as  Problem \ref{prob:learning_few}  below.

\begin{problem}[Learning few examples]\label{prob:learning_few} Consider a classifier $F$ defined by (\ref{eq:classifier_general}), and let $\mathcal{U}_{\mathrm{new}}=\{\bfu_1,\dots,\bfu_k\}$, $k\in\Natural$, $\bfu_i\in\mathcal{U}_{\mathrm{new}}$, be a given finite set to be learned by $F$. Let $\ell_{\mathrm{new}}\in\mathcal{L}$  be a  label associated with the new set $\mathcal{U}_{\mathrm{new}}$. Let $p_e$ be a given positive number in the interval $(0,1]$ determining the quality of learning. 

Find an algorithm $\mathcal{A}(\mathcal{U}_{\mathrm{new}})$ producing a function $g^\ast: \mathcal{X}\rightarrow \mathcal{L}$ such that
\begin{equation}\label{eq:learining_few_1}
F[g^\ast\circ f(\bfu)] = \ell_{\mathrm{new}} \ \mbox{for all} \ \bfu\in\mathcal{U}_{\mathrm{new}}
\end{equation}
and
\begin{equation}\label{eq:learining_few_2}
P\big(F[g^\ast\circ f(\bfu)] = F[g\circ f(\bfu)]\big)\geq p_e
\end{equation}
for $\bfu$ drawn from the distribution $P_u$.
\end{problem}

\subsubsection{Learning from an arbitrary finite number of examples} 

Let us now consider a different version of the problem where the system is to {\it learn a new class} from few examples. The key difference here from the case considered in Problem \ref{prob:learning_few} is that we will no longer require that all new examples are memorised. Instead, we will request that {\it all } elements of the new class are assigned a correct label with some a-priori defined probability. At the same time, we will request that performance of the classifier on elements from other classes does not drop below a given predefined and acceptable level. 

Extending our earlier conventions, we will suppose that the new class can be described by a corresponding probability distribution $P_{\mathrm{new}}$ and will be associated with a new label $\ell_{\mathrm{new}}$. Formal statement of this task is provided in Problem \ref{prob:learning_from_few}.

\begin{problem}[Learning from few examples]\label{prob:learning_from_few} Consider a classifier $F$ defined by (\ref{eq:classifier_general}), and let $\mathcal{U}_{\mathrm{new}}=\{\bfu_1,\dots,\bfu_k\}$, $k\in\Natural$, $\bfu_i\in\mathcal{U}_{\mathrm{new}}$, be a  finite  independent and identically distributed (i.i.d.) sample from a distribution $P_{\mathrm{new}}$, and  $\ell_{\mathrm{new}}\in\mathcal{L}$ be a corresponding new label to be associated with the elements drawn from $P_{\mathrm{new}}$. Let $p_e, p_n$ be given positive numbers in the interval $(0,1]$ determining the quality of learning. 

Find an algorithm $\mathcal{A}(\mathcal{U}_{\mathrm{new}})$ producing a function $g^\ast: \mathcal{X}\rightarrow \mathcal{L}$ such that
\begin{equation}\label{eq:learining_from_few_1}
P\big(F[g^\ast\circ f(\bfu)] = \ell_{\mathrm{new}} \big) \geq p_n
\end{equation}
for $\bfu$ drawn from $P_{\mathrm{new}}$, and
\begin{equation}\label{eq:learining_from_few_2}
P\big(F[g^\ast\circ f(\bfu)] = F[g\circ f(\bfu)]\big)\geq p_e
\end{equation}
for $\bfu$ drawn from the distribution $P_u$.
\end{problem}

\begin{rem} Note that Problems \ref{prob:learning_few}, \ref{prob:learning_from_few} do not rely upon standard relationships between expected and empirical risks to characterise generalisation and learning. Instead, they impose stronger requirements: lower bounds on probabilities of success. 

These stronger requirements have clear practical benefits in terms of understanding limitations and capabilities of few-shot learning algorithms $\mathcal{A}$. Potential downsides, however, are that knowledge of some general properties of the data distributions (support, non-degeneracy, etc) may  be needed to guarantee that these stronger requirements could be met.
\end{rem}


\subsection{Assumptions}\label{sec:assumptions}

In  agreement with existing literature on few-shot learning \cite{vinyals2016matching}, \cite{snell2017prototypical}, we will primarily be dealing with  representations $\bfx=f(\bfu)$ of inputs $\bfu\in\mathcal{U}$ in the system's latent space $\mathcal{X}$ as opposed to working directly with $\mathcal{U}$ (see  Fig. \ref{fig:scheme_classifier} for a diagram of the workflow).  We will hence assume that the distributions $P_u$, $P_{\ell|u}$, $P_{\mathrm{new}}$, and the function $f$ in (\ref{eq:classifier_general})  -- (\ref{eq:learining_from_few_2}) induce their corresponding distributions $P_x$, $P_{\ell|x}$, $P_{\mathrm{new},x}$ in the system's latent space $\mathcal{X}$.

We will further assume that distributions $P_x$, $P_{\mathrm{new},x}$ are supported on some balls in $\Real^n$ and admit probability density functions satisfying some non-degeneracy constraints. Formally these requirements are formulated in Assumptions \ref{assume:P}, \ref{assume:P_new}.

\begin{assume}\label{assume:P} The probability density function $p_x$ associated with  $P_x$ exists, is defined on the unit ball $\mathbb{B}_n$, and there exist constants $C_x, r > 0$  such that
\[
p_x(\bfx) \leq \frac{C_x}{V_n(\mathbb{B}_n)}   r^n.
\]
\end{assume}

\begin{assume}\label{assume:P_new} The probability density function $p_{\mathrm{new},x}$ associated with  $P_{\mathrm{new},x}$ exists, is defined on a ball $\mathbb{B}_n(v,\bfc)$, and there exist constants $C_{\mathrm{new},x}, \rho > 0$  such that
\[
p_{\mathrm{new},x}(\bfx) \leq \frac{C_{\mathrm{new},x} }{V_n(\mathbb{B}_n)} \rho^n.
\]
\end{assume}

In the next section we present main theoretical findings and  quantifying success of few- and one-shot learning schemes. These results join together various ideas presented in earlier works  \cite{tyukin2017high},  \cite{GorTyu:2016}, \cite{tyukin2017knowledge}, \cite{gorban2018correction}, \cite{GorMakTyu:2018}, \cite{gorban2020high}, and reveal intrinsic links between data dimensionality, partial knowledge about data models, and generalisation bounds. 

\section{Main Results}\label{sec:results}

\subsection{Learning an arbitrary finite number of examples}

Our first result concerns  Problem \ref{prob:learning_few} and is formally expressed in Theorem \ref{thm:learning_few_examples} below

\begin{thm}\label{thm:learning_few_examples}[Learning few examples] Consider a classifier $F$ defined by (\ref{eq:classifier_general}), and let $\mathcal{U}_{\mathrm{new}}=\{\bfu_1,\dots,\bfu_k\}$, $k\in\Natural$, $\bfu_i\in\mathcal{U}_{\mathrm{new}}$, be a  finite set, and  $\ell_{\mathrm{new}}\in\mathcal{L}$ be a corresponding new label to be associated with the elements from this new set.

Let $\mathcal{Y}=\{\bfx_1,\dots,\bfx_k\}$, $\bfx_i=f(\bfu_i)$, $i=1,\dots,k$ be a representation of the set $\mathcal{U}_{\mathrm{new}}$ in the classifier's latent space, 
\[
\bar{\bfx}=\frac{1}{k} \sum_{i} \bfx_i,
\]
be the empirical mean of the representation with
\[
(\bar{\bfx},\bfx_i)\geq 0 \ \mbox{for all} \ \bfx_i\in\mathcal{Y},
\]
and let Assumption \ref{assume:P} hold.

Then the map
\begin{equation}\label{eq:learning_few_algorithm}
g^\ast: \ g^\ast(\bfx)=\left\{\begin{array}{ll}
						   \ell_{\mathrm{new}} , & \mbox{if} \ \left(\frac{\bar{\bfx}}{\|\bar{\bfx}\|}, \bfx \right) - \theta \geq 0\\
						   g(\bfx) , & \mbox{otherwise}		
						   \end{array}\right.
\end{equation}
parameterised by
\[
\theta=\min_{i\in \{1,\dots,k\}} \left\{\left(\frac{\bar\bfx}{\|\bar\bfx\|},\bfx_i \right)\right\}
\]
is a solution of Problem \ref{prob:learning_few} with 
\begin{equation}\label{eq:thm:learning_few:bound}
p_e= 1 - \frac{C_x}{2} \left[r \left(1-\theta^2\right)^{1/2}\right]^n.
\end{equation}
\end{thm} 
{\it Proof of Theorem \ref{thm:learning_few_examples}}. According to the definition of the map $g^\ast$ and the fact that $\bfx_i=f(\bfu_i)$, 
\[
\left(\frac{\bar{\bfx}}{\|\bar{\bfx}\|}, \bfx_i \right) - \theta\geq 0
\]
and as a result 
\[
F[g^\ast \circ f(\bfu_i)]=\ell_{\mathrm{new}}
\]
for any $\bfu_i\in \mathcal{U}_{\mathrm{new}}$. 

Let $\bfu$ be drawn from the distribution $P_u$.  This vector has a latent representation $\bfx=f(\bfu)$ and a corresponding induced distribution $P_x$ satisfying Assumption \ref{assume:P}. Let 
\[
\mathcal{C}_{n}(\bfz,\theta)=\left\{\bfx\in\mathbb{B}_{n} \left| \ \left(\frac{\bfz}{\|\bfz\|}, \bfx \right) - \theta \geq  0 \right. \right\}.
\]
Then
\[
\begin{split}
& P(F[g^\ast \circ f(\bfu) ]= \ell_{\mathrm{new}}) = \int_{\mathcal{C}_{n}(\bar\bfx,\theta)} p_x(\bfx)d \bfx \\
\leq &  \frac{C_x}{V_n(\mathbb{B}_n)} r^n  \int_{\mathcal{C}_{n}(\bar\bfx,\theta)} d\bfx \leq  \frac{C_x}{2} r^n  [ (1-\theta^2)^{1/2}]^n.
\end{split}
\]
The statement now follows. $\square$

\begin{rem} Note that if $r(1-\theta^2)^{1/2}<1$ then the bound $p_e$ approaches $1$ exponentially fast as $n$ grows. This implies that learning a single or few examples can be efficiently accomplished by an exceptionally simple map (\ref{eq:learning_few_algorithm}). 

Performance of this learning scheme depends on the values of $r$ and $\theta$. The closer the value of $\theta$ is to $1$, however, the broader the range of $r$ for which solution (\ref{eq:learning_few_algorithm}) of Problem \ref{prob:learning_few} is appropriate.
\end{rem}

\begin{rem}  One can easily verify that few-shot learning scheme (\ref{eq:learning_few_algorithm}) assigns the label $\ell_{\mathrm{new}}$ to all convex combinations of $\bfu_i\in\mathcal{U}_{\mathrm{new}}$. In this respect, the entire convex hull of  $\mathcal{U}_{\mathrm{new}}$ is learnt by (\ref{eq:learning_few_algorithm}).  
\end{rem}

In the next subsection we will show that, under appropriate assumptions, learning schemes which are very similar to (\ref{eq:learning_few_algorithm}) have a capacity to generalise beyond  finite sets and their convex hulls from just few examples. 

\subsection{Learning from few examples}

Let us now turn attention to Problem \ref{prob:learning_from_few}. Our main theoretical statement specifying a simple solution of this problem is presented in Theorem \ref{thm:learning_from_few_examples}.  Similarly to Theorem \ref{thm:learning_few_examples}, we show that performance of the proposed scheme to learn from $k$ examples is closely related to 1) dimension $n$ of the classifier's latent space and 2) non-degeneracy of probability distributions of the inputs' representations in that space.

The theorem is largely based on Lemmas \ref{lem:orthogonality}, \ref{lem:centering} which we present below.

\begin{lem}\label{lem:orthogonality} Let $\mathcal{Y}=\{\bfx_1,\bfx_2,\dots,\bfx_k\}$ be a set of $k$ i.i.d. random vectors drawn from a distribution satisfying Assumption \ref{assume:P_new}, and let $\delta,\varepsilon\in(0,1)$. Consider event $A_1$:
\begin{equation}\label{eq:event_1}
A_1:  \ | (\bfx_i-\bfc,\bfx_j-\bfc)| \leq  {\delta v}, \ \forall \  i\neq j
\end{equation}
and event $A_2$:
\begin{equation}\label{eq:event_2}
A_2:  \  \|\bfx_i-\bfc\|\geq (1-\varepsilon)v \ \forall \ i.
\end{equation}
Then 
\begin{equation}\label{eq:lem:orthogonality:statement:1}
P(  A_1 ) \geq 1 - C_{\mathrm{new}} \frac{k (k-1)}{2} \left[\rho v (1-\delta^2)^{1/2}\right]^n,
\end{equation}
and
\begin{equation}\label{eq:lem:orthogonality:statement:2}
\begin{split}
&P\left(  A_1   \wedge  A_2 \right) \geq \\
& 1 - C_{\mathrm{new}} k[\rho v (1-\varepsilon)]^n - C_{\mathrm{new}} \frac{k (k-1)}{2} \left[\rho v (1-\delta^2)^{1/2}\right]^n.
\end{split}
\end{equation}
\end{lem}
{\it Proof of Lemma  \ref{lem:orthogonality}}. Let us denote $\tilde{\bfx}_i = \bfx_i-\bfc$. Consider events
\[
\begin{split}
E_1: &  \ \left|\left(\frac{\tilde{\bfx}_1}{\|\tilde{\bfx}_1\|},\tilde{\bfx}_2\right)\right| > \delta \\
E_2: &  \ \left[\left|\left(\frac{\tilde{\bfx}_1}{\|\tilde{\bfx}_1\|},\tilde{\bfx}_3\right)\right|>\delta\right] \vee \left[\left|\left(\frac{\tilde{\bfx}_2}{\|\tilde{\bfx}_2\|},\tilde{\bfx}_3\right)\right|  > \delta\right] \\
\vdots & \ \ \ \ \ \ \ \ \  \ \ \ \ \  \vdots \\
E_{k-1}: &  \ \left[\left|\left(\frac{\tilde{\bfx}_1}{\|\tilde{\bfx}_1\|},\tilde{\bfx}_{k}\right)\right| > \delta \right] \vee \cdots \vee \left[\left|\left(\frac{\tilde{\bfx}_{k-1}}{\|\tilde{\bfx}_{k-1}\|}, \tilde{\bfx}_k\right)\right|  > \delta\right]\\
& \\
B_1: &  \ \|\tilde{\bfx}_1\| < (1-\varepsilon)v \\
\vdots & \ \ \ \ \ \ \ \ \  \ \ \ \ \  \vdots \\
B_k: &  \ \|\tilde{\bfx}_k\| < (1-\varepsilon)v 
\end{split}
\]
Let 
\[
\mathcal{C}_{n}(\bfz,\bfc,v,\delta)=\left\{\bfx\in\mathbb{B}_{n}(v,\bfc) \left| \ \left(\frac{\bfz}{\|\bfz\|}, \bfx-\bfc \right)>\delta \right. \right\}.
\]
According to Assumption \ref{assume:P_new} and the fact that $\bfx_1$ and $\bfx_2$ are drawn independently from the same distribution, the probability that event $E_1$ occurs can be bounded from above as
\begin{equation}\label{eq:lem:orthogonality:1}
\begin{split}
P(E_1)=& \int_{\mathcal{C}_n(\tilde{\bfx}_1,\bfc,v,\delta)} p_{\mathrm{new},x}(\bfx)d\bfx \\
& \ \ \ \ \ \ \ \ \ \ \ \ +  \int_{\mathcal{C}_n(-\tilde{\bfx}_1,\bfc,v,\delta)} p_{\mathrm{new},x}(\bfx)d\bfx \\
< &  \frac{C_{\mathrm{new},x} \rho^n}{V_n(\mathbb{B}_n)} \int_{\mathcal{C}_n(\tilde{\bfx}_1,\bfc,v,\delta)} d\bfx \\
& \ \ \ \ \ \ \ \ \ \ \ \ + \frac{C_{\mathrm{new},x} \rho^n}{V_n(\mathbb{B}_n)} \int_{\mathcal{C}_n(-\tilde{\bfx}_1,\bfc,v,\delta)} d\bfx \\
 = & \frac{C_{\mathrm{new},x} \rho^n}{V_n(\mathbb{B}_n)} 2 V_n (\mathcal{C}_{n}(\tilde{\bfx}_1,\bfc,v,\delta)).
\end{split}
\end{equation}
Observe that
\begin{equation}\label{eq:lem:orthogonality:2}
2 V_n \big(\mathcal{C}_{n}(\tilde{\bfx}_1,\bfc,v,\delta)\big) \leq V_n(\mathbb{B}_n) [v(1-\delta^2)^{1/2}]^n.
\end{equation}
Combining (\ref{eq:lem:orthogonality:1}), (\ref{eq:lem:orthogonality:2}) we obtain:
\begin{equation}\label{eq:lem:orthogonality:3}
P(E_1) <  C_{\mathrm{new},x}   [\rho v(1-\delta^2)^{1/2}]^n.
\end{equation}
Recall that for any events $A_1,\dots,A_k$ the following probability union bound holds true (also known as the Boole's inequality):
\begin{equation}\label{eq:union_bound}
P(\bigcup_{i=1}^k A_i)\leq \sum_{i=1}^k P(A_i).
\end{equation}
Hence
\[
\begin{split}
P(E_2)=&P \left(\left[\left|\left(\frac{\tilde{\bfx}_1}{\|\tilde{\bfx}_1\|},\tilde{\bfx}_3\right)\right|>\delta \right] \vee \left[\left|\left(\frac{\tilde{\bfx}_2}{\|\tilde{\bfx}_2\|},\tilde{\bfx}_3\right)\right|  > \delta\right]\right)\\
\leq &\sum_{i=1}^2 P \left(\left|\left(\frac{\tilde{\bfx}_i}{\|\tilde{\bfx}_i\|},\tilde{\bfx}_3\right)\right|>\delta\right).
\end{split}
\]
Using the same argument as in (\ref{eq:lem:orthogonality:1})--(\ref{eq:lem:orthogonality:3}) we can derive that
\[
P(E_2) < 2 C_{\mathrm{new},x}   [\rho v(1-\delta^2)^{1/2}]^n,
\]
and  that
\begin{equation}\label{eq:lem:orthogonality:4}
P(E_i) < i \cdot C_{\mathrm{new},x}   [\rho v(1-\delta^2)^{1/2}]^n \ \mbox{for all} \ i=1,\dots,k.
\end{equation}

Consider now events $B_1,\dots B_k$ and evalutate $P(B_i)$, $i=1,\dots,k$:
\begin{eqnarray}\label{eq:lem:orthogonality:5}
 && P(B_i)=\int_{\mathbb{B}_n(v(1-\varepsilon),\bfc)} p_{\mathrm{new},x}(\bfx)d\bfx \\
& &\leq  \frac{C_{\mathrm{new},x}\rho^n}{V_n(\mathbb{B}_n)}  \int_{\mathbb{B}_n(v(1-\varepsilon),\bfc)} d\bfx \nonumber \\
& &= C_{\mathrm{new},x}\rho^n \frac{ V_n(\mathbb{B}_n(v(1-\varepsilon),\bfc)}{V_n(\mathbb{B}_n)}  =  C_{\mathrm{new},x}[\rho v (1-\varepsilon)]^n. \nonumber
\end{eqnarray}

Recall that, for any sets $A_1,\dots,A_d$,  De Morgan's law states that:
\[
{\bigwedge_{i=1}^d A_i} = \mbox{not} \ \left({\bigvee_{i=1}^d (\mbox{not} \ {A_i}) }\right).
\]
Therefore
\[
\begin{split}
&P(A_1 \land  A_2 \land \cdots \land A_d)=\\
&\ \ \ \ \ \ \ \ \ \ \ 1-P( (\mbox{not} \  A_1)\lor (\mbox{not} \ A_2) \lor \cdots (\mbox{not}  \ A_d)).
\end{split}
\]
Using the union bound rule (\ref{eq:union_bound}), one can derive that
\begin{equation}\label{eq:prob_bound}
P(A_1 \land A_2 \land \cdots \land A_d)\geq 1 - \sum_{i=1}^d P(\mbox{not} \ A_i).
\end{equation}

To complete the lemma, consider  events
\[
\begin{split}
\mbox{not}  \ E_1: &  \ \left|\left(\frac{\tilde{\bfx}_1}{\|\tilde{\bfx}_1\|},\tilde{\bfx}_2\right)\right| \leq  \delta \\
\mbox{not}  \ E_2: &  \ \left[\left|\left(\frac{\tilde{\bfx}_1}{\|\tilde{\bfx}_1\|},\tilde{\bfx}_3\right)\right| \leq \delta\right]  \wedge \left[\left|\left(\frac{\tilde{\bfx}_2}{\|\tilde{\bfx}_2\|},\tilde{\bfx}_3\right)\right|  \leq  \delta\right] \\
\vdots & \ \ \ \ \ \ \ \ \  \ \ \ \ \  \vdots \\
\mbox{not} \ E_{k-1}: &  \ \left[ \left|\left(\frac{\tilde{\bfx}_1}{\|\tilde{\bfx}_1\|},\tilde{\bfx}_{k}\right)\right|\leq \delta\right]  \wedge \cdots\\
& \ \ \ \ \ \ \ \ \ \ \ \ \ \ \  \wedge \left[\left|\left(\frac{\tilde{\bfx}_{k-1}}{\|\tilde{\bfx}_{k-1}\|}, \tilde{\bfx}_k\right)\right|  \leq  \delta\right]\\
\mbox{not}  \ B_1: &  \ \|\tilde{\bfx}_1\| \geq  (1-\varepsilon)v \\
\vdots & \ \ \ \ \ \ \ \ \  \ \ \ \ \  \vdots \\
\mbox{not}  \ B_k: &  \ \|\tilde{\bfx}_k\| \geq  (1-\varepsilon)v 
\end{split}
\]
Given that $\bfx_i\in\mathbb{B}_n(v,\bfc)$, we have that $\|\tilde{\bfx_i}\|\leq v$. It is hence clear that the event
$[\mbox{not} \ E_1 \wedge \cdots \wedge  \mbox{not} \ E_{k-1}]$ is contained in the event $A_1$ defined by (\ref{eq:event_1}) in the sense that  any $\bfx_1,\dots, \bfx_k\in\mathbb{B}_n(v,\bfc)$  for which 
\[
\mbox{not} \ E_1 \wedge \cdots \wedge  \mbox{not} \ E_{k-1}
\]
holds true must necessarily satisfy (\ref{eq:event_1}). 

Therefore, according to (\ref{eq:event_2}), (\ref{eq:prob_bound}), we can write:
\[
P(A_1)\geq   P(\mbox{not} \ E_1 \wedge\dots\wedge \mbox{not} \ E_{k-1}) \geq  1 - \sum_{i=1}^{k-1} P(E_i),
\]
and
\[
\begin{split}
& P(A_1 \wedge A_2)= P (A_1 \wedge \mbox{not}  \ B_1 \wedge \cdots \mbox{not}  \ B_k)\\ 
&\geq  P(\mbox{not} \ E_1 \wedge\dots\wedge \mbox{not} \ E_{k-1} \wedge  \mbox{not} \ B_1 \wedge \dots \wedge \mbox{not} \ B_k \cdots)\\
& \geq 1 - \sum_{i=1}^{k-1} P(E_i) - \sum_{i=1}^{k} P(B_i).
\end{split}
\]
Substituting (\ref{eq:lem:orthogonality:4}), (\ref{eq:lem:orthogonality:5}) into the latter expressions one can now conclude that the lemma holds true. $\square$

\begin{rem}  Lemma \ref{lem:orthogonality} reveals, in a general setting,  the typicality of large ``almost'' or quasi- orthogonal bases in high-dimension  (cf. \cite{GorTyu:2016}). Indeed, according to (\ref{eq:lem:orthogonality:statement:2}), if $\rho v (1-\varepsilon)<1$, $\rho v (1-\delta^2)^{1/2}<1$ then
\[
\begin{split}
&|\cos(\bfx_i-\bfc,\bfx_j-\bfc)|=\frac{|(\bfx_i-\bfc,\bfx_j-\bfc)|}{\|\bfx_i-\bfc\| \|\bfx_j-\bfc\|} \\
&\leq \frac{\delta v}{v^2(1-\varepsilon)^2} = \frac{\delta }{v (1-\varepsilon)^2}
\end{split}
\]
with probability close to $1$ if $n$ is sufficiently large. Earlier works \cite{Kurkova} (see also \cite{kainen2020quasiorthogonal}, \cite{kainen1997utilizing})  showed that large ($k\gg n$) quasi-orthogonal bases exist. Here we follow our earlier results \cite{GorTyu:2016} and prove that almost orthogonal corteges of vectors whose cardinalily $k$ grows exponentially with dimension $n$ are {\it typical} in high dimension.
\end{rem}

Our next result, Lemma \ref{lem:centering} shows how this almost or quasi-orthogonality property can be used to estimate centroids of data clusters in high-dimensional datasets from few observations. 

\begin{lem}\label{lem:centering}  Let $\mathcal{Y}=\{\bfx_1,\bfx_2,\dots,\bfx_k\}$ be a set of $k$ i.i.d. random vectors drawn from a distribution satisfying Assumption \ref{assume:P_new}, and let $\delta,\varepsilon\in(0,1)$. Let
\[
\bar\bfx  = \frac{1}{k} \sum_{i=1}^k \bfx_i
\]
be the empirical mean of the sample. 

Then
\begin{equation}\label{eq:lem:centering:1}
\begin{split}
& P\left(L(k,\delta,\varepsilon)  \leq \|\bar\bfx-\bfc\|^2 \leq U(k,\delta) \right)\\
&\ \ \ \ \ \ \ \ \ \ \ \ \ \ \ \ \ \geq 1 - R_{\varepsilon,\delta}(n,k,v,\rho,\delta,\varepsilon),
\end{split}
\end{equation}
and
\begin{equation}\label{eq:lem:centering:2}
 P\left(\|\bar\bfx-\bfc\|^2 \leq U(k,\delta) \right) \geq 1 - R_{\delta}(n,k,v,\rho,\delta),
\end{equation}
where
\[
\begin{split}
U(k,\delta)=\frac{v^2}{k} + \frac{k-1}{k} v\delta\\
L(k,\delta,\varepsilon)=\frac{(1-\varepsilon)^2 v^2}{k} - \frac{k-1}{k} v\delta
\end{split}
\]
and
\[
\begin{split}
R_{\varepsilon,\delta}(n,k,v,\delta,\rho,\varepsilon)= & C_{\mathrm{new}} k[\rho v (1-\varepsilon)]^n \\
& + C_{\mathrm{new}} \frac{k (k-1)}{2} \left[\rho v (1-\delta^2)^{1/2}\right]^n, \\
R_{\delta}(n,k,v,\rho,\delta)= &  C_{\mathrm{new}} \frac{k (k-1)}{2} \left[\rho v (1-\delta^2)^{1/2}\right]^n.
\end{split}
\]
\end{lem}
{\it Proof of Lemma \ref{lem:centering}}. The Lemma is essentially contained in Lemma \ref{lem:orthogonality}. Indeed, consider
\[
\begin{split}
& \|\bar\bfx-\bfc\|^2 = (\bar\bfx-\bfc, \bar\bfx-\bfc)= \left(\frac{1}{k}\sum_{i=1}^k \bfx_i - \bfc, \frac{1}{k}\sum_{i=1}^k \bfx_i - \bfc \right) \\
&=\frac{1}{k^2} \sum_{i=1}^k \|\bfx_i-\bfc\|^2 + \frac{1}{k^2} \sum_{i\neq j}  (\bfx_i - \bfc,\bfx_j - \bfc ).
\end{split}
\]
According to Lemma \ref{lem:orthogonality} (statement (\ref{eq:lem:orthogonality:statement:1})), the term
\[
\left| \frac{1}{k^2} \sum_{i\neq j}  (\bfx_i - \bfc,\bfx_j - \bfc )\right| \leq \frac{k-1}{k} v \delta
\]
with probability $1-R_{\delta}(n,k,v,\delta)$. This, together with the fact that $\|\bfx_i-\bfc\|\leq v $ for all $i=1,\dots,k$, prove (\ref{eq:lem:centering:1}). 
Similarly, statement (\ref{eq:lem:orthogonality:statement:2}) of Lemma \ref{lem:orthogonality} implies now that bound (\ref{eq:lem:centering:2}) holds true too.
$\square$

\begin{thm}\label{thm:learning_from_few_examples}[Learning from few examples] Consider a classifier $F$ defined by (\ref{eq:classifier_general}), and let $\mathcal{U}_{\mathrm{new}}=\{\bfu_1,\dots,\bfu_k\}$, $k\in\Natural$, $\bfu_i\in\mathcal{U}_{\mathrm{new}}$, be a  finite  independent and identically distributed (i.i.d.) sample from a distribution $P_{\mathrm{new}}$, and  $\ell_{\mathrm{new}}\in\mathcal{L}$ be a corresponding new label to be associated with the elements drawn from $P_{\mathrm{new}}$.

Let $\mathcal{Y}=\{\bfx_1,\dots,\bfx_k\}$, $\bfx_i=f(\bfu_i)$, $i=1,\dots,k$ be a representation of the sample $\mathcal{U}_{\mathrm{new}}$ in the classifier's latent space, 
\[
\bar{\bfx}=\frac{1}{k} \sum_{i} \bfx_i,
\]
be the empirical mean of the representation, and let Assumption \ref{assume:P_new} hold. Finally, let $\delta\in(0,1)$ be a number satisfying
\[
\Delta=\|\bar{\bfx}\|-\left(\frac{v^2}{k}+\frac{k-1}{k}v\delta\right)^{1/2} >0.
\]

Then the map
\begin{equation}\label{eq:learning_from_few_algorithm}
g^\ast: \ g^\ast(\bfx)=\left\{\begin{array}{ll}
						   \ell_{\mathrm{new}} , & \mbox{if} \ \left(\frac{\bar{\bfx}}{\|\bar{\bfx}\|}, \bfx \right) - \theta \geq 0\\
						   g(\bfx) , & \mbox{otherwise}		
						   \end{array}\right.
\end{equation}
parameterised by
\[
\theta\in \left[\max\{\Delta  - v, 0,\}, \Delta \right]
\]
is a solution of Problem \ref{prob:learning_from_few} with 
\begin{equation}\label{eq:thm:learning_from_few:bound}
\begin{split}
p_n=& \left(1 - \frac{C_{\mathrm{new},x}}{2} \left[\rho \left(v^2-(\Delta-\theta)^2\right)^{1/2}\right]^n\right) \times \\
 & \ \ \ \ \ \ \ \ \ \left(1- C_{\mathrm{new}} \frac{k (k-1)}{2} \left[\rho v (1-\delta^2)^{1/2}\right]^n\right),\\
p_e=& 1 - \frac{C_x}{2} \left[r \left(1-\theta^2\right)^{1/2}\right]^n.
\end{split}
\end{equation}
\end{thm}
{\it Proof of Theorem \ref{thm:learning_from_few_examples}}. According to Lemma \ref{lem:centering}, the probability that the centre $\bfc$ is within 
$\left(\frac{v^2}{k}+\frac{k-1}{k}v\delta\right)^{1/2}$ from the empirical mean $\bar\bfx$ is at least
\[
\left(1- C_{\mathrm{new}} \frac{k (k-1)}{2} \left[\rho v (1-\delta^2)^{1/2}\right]^n\right).
\]

\begin{figure}
\centering
\includegraphics[width=\columnwidth]{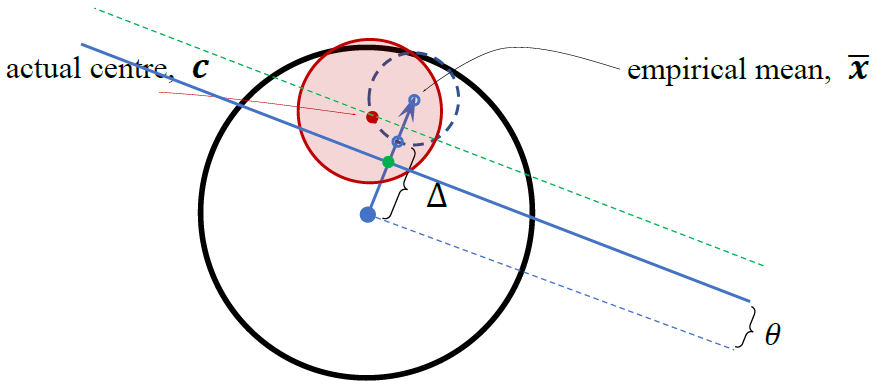}
\caption{Illustration to the proof of Theorem \ref{thm:learning_from_few_examples}. Pink filled circle shows $\mathbb{B}_n(v,\bfc)$, blue dashed disc shows the domain where the true centre $\bfc$ is located (with high probability), and blue solid line shows the hyperplane $\left(\frac{\bar{\bfx}}{\|\bar{\bfx}\|}, \bfx \right) - \theta=0$. }\label{fig:proof_sketch}
\end{figure}

Suppose that the above event occurs. This implies that the hyperplane 
\[
\bfx: \left(\frac{\bar{\bfx}}{\|\bar{\bfx}\|}, \bfx \right) - \theta = 0
\]
is at least $\Delta-\theta$ away from the hyperplane with the same normal, $\frac{\bar{\bfx}}{\|\bar{\bfx}\|}$, and which is passing through the centre $\bfc$ of the ball $\mathbb{B}_n(v,\bfc)$ (see Fig. \ref{fig:proof_sketch}). 

The probability that an element drawn from the distribution $P_{\mathrm{new}}$ would have a representation $\bfx$ for which $\left(\frac{\bar{\bfx}}{\|\bar{\bfx}\|}, \bfx \right) - \theta >0$ is hence at least
\[
1 - \frac{C_{\mathrm{new},x}}{2} \left[\rho \left(v^2-(\Delta-\theta)^2\right)^{1/2}\right]^n.
\]
This justifies the expression for $p_n$ in (\ref{eq:thm:learning_from_few:bound}). 

Similarly, the probability that an element drawn from the distribution  $P_u$ would be assigned a label $\ell_{\mathrm{new}}$ is at most
\[
\frac{C_x}{2} \left[r \left(1-\theta^2\right)^{1/2}\right]^n.
\]
Hence the expression for $p_e$ follows. $\square$

\begin{rem} According to Theorem \ref{thm:learning_from_few_examples} and similar to the case covered by Theorem \ref{thm:learning_few_examples}, under appropriate and reasonable assumptions, the probabilities of success in the task of learning from few examples  approach $1$ exponentially fast as $n$ grows.
\end{rem}

\subsection{Discussion}

Having introduced our main theoretical results, let us now briefly relate these results to existing literature on few-shot learning and outline future potential direction s.

\subsubsection{Matching  and prototypical networks}

Theorems \ref{thm:learning_few_examples}, \ref{thm:learning_from_few_examples} and few-shot learning algorithms (\ref{eq:learning_few_algorithm}), (\ref{eq:learning_from_few_algorithm}), which these theorems relate to, show striking similarity to approaches presented and empirically studied in \cite{vinyals2016matching}, \cite{snell2017prototypical}. In the case  of one-shot learning \cite{vinyals2016matching}, Theorem \ref{thm:learning_few_examples} with $k=1$ applies, whereas in the case of few-shot learning, \cite{snell2017prototypical}, Theorem \ref{thm:learning_from_few_examples} could be more appropriate for explaining and interpreting why few-shot learning works. 


\subsubsection{Object models and the challenge of generalisation} 

Our results show that significant understanding and insights into why and when large-scale and highly expressive AI systems, including deep neural networks, can generalise well from just few examples can be gained if some loose assumptions are introduced on the data models. In our case, these assumptions, are that 1) the probability distributions of objects' representations in the system's latent space are supported on some balls (or ellipsoids, subject to a coordinate transformation), and 2) these probability distributions are not degenerate in the sense of Assumptions \ref{assume:P}, \ref{assume:P_new}. Going forward, one can consider further straightforward generalisations in which the objects are modeled by mixtures of these models. These generalisations, are however, beyond the scope of the current work.

In addition, our current work, by focusing on what can and what cannot be learned from few examples in randomised settings, provides insights into  why  stochastic configuration networks may be so successful in practice \cite{wang2017stochastic}, \cite{huang2019stochastic}: practically relevant functions we are interested to learn may have a ``compact'' structure, and the process of stochastic configuration could be viewed as an efficient mechanism that is capable to learn this structure from data step-by-step.

\subsubsection{Learning to learn}
 
In addition to explaining why few-shot learning models work and why large-scale deep learning models may generalise so well, our present work {\it presents high-level training requirements} for a model that is trained to learn from few examples. These requirements are specified in Assumptions \ref{assume:P} and \ref{assume:P_new}. If a network is trained so that object representations in its latent space satisfy Assumptions \ref{assume:P} and \ref{assume:P_new} with appropriate relevant constants then Theorems \ref{thm:learning_few_examples}, \ref{thm:learning_from_few_examples} guarantee that such models can indeed learn from mere few or single examples. Importantly, training of networks to satisfy  Assumptions \ref{assume:P} and \ref{assume:P_new} can be posed within the standard empirical risk minimisation framework. A very similar approach has been pursued in \cite{snell2017prototypical}, \cite{vinyals2016matching}, albeit heuristically.   

\section{Conclusion}\label{sec:conclusion}

This work presents a formal treatment of the challenges of few-shot and one-shot learning and generalisation in large-scale modern AI models. We provided formal statements of these learning  problems and  showed that high dimensionality and geometry of objects' representations in the systems' latent spaces along with some non-degeneracy conditions are key determinants explaining when and why such learning is possible. 

Our results suggest that neural  networks' generalisation capabilities are intrinsically linked with internal regularities in the data sets and also with representations of these regularities  in the networks' latent spaces.  The results reveal an important characteristic of this important regularity: if an object  has a ``compact'' representation in the network's latent space then such object can be learned from just few or even single example. Absence of  such compact representations may require exponentially large training samples to learn from.

\bibliographystyle{IEEEtran}
\bibliography{IEEEabrv,few_shot_refs}

\end{document}